\documentclass[runningheads]{llncs}
\usepackage{graphicx}
\usepackage{amsmath,amssymb} 
\usepackage{color}

\begin{document}
\title{NeXtVLAD: An Efficient Neural Network to Aggregate Frame-level Features for Large-scale Video Classification}

\titlerunning{NeXtVLAD}
%
\author{Rongcheng Lin, Jing Xiao, Jianping Fan}
%
\authorrunning{R. Lin, J. Xiao, J. Fan}
%

\institute{University of North Carolina at Charlotte\\
\email{\{rlin4, xiao, jpfan\}@uncc.edu}}
\maketitle              
\begin{abstract}
This paper introduces a fast and efficient network architecture, NeXtVLAD, to aggregate frame-level features into a compact feature vector for large-scale video classification. Briefly speaking, the basic idea is to decompose a high-dimensional feature into a group of relatively low-dimensional vectors with attention before applying NetVLAD aggregation over time. This NeXtVLAD approach turns out to be both effective and parameter efficient in aggregating temporal information. In the 2nd Youtube-8M video understanding challenge, a single NeXtVLAD model with less than 80M parameters achieves a GAP score of 0.87846 in private leaderboard. A mixture of 3 NeXtVLAD models results in 0.88722, which is ranked 3rd over 394 teams. The code is publicly available at \url{https://github.com/linrongc/youtube-8m}.

\keywords{Neural Network \and VLAD \and Video Classification \and Youtube8M}
\end{abstract}
\section{Introduction}
The prevalence of digital cameras and smart phones exponentially increases the number of videos, which are then uploaded, watched and shared through internet. Automatic video content classification has become a critical and challenging problem in many real world applications, including video-based search, recommendation and intelligent robots etc. To accelerate the pace of research in video content analysis, Google AI launched the second Youtube-8M video understanding challenge, aiming to learn more compact video representation under limited budget constraints. Because of both unprecedent scale and diversity of Youtube-8M dataset\cite{45619}, they also provided the frame-level visual and audio features which are extracted by pre-trained convolutional neural networks (CNNs). The main challenge is how to aggregate such pre-extracted features into a compact video-level representation effectively and efficiently.

NetVLAD, which was developed to aggregate spatial representation for the task of place recognition\cite{Arandjelovic16}, was found to be more effective and faster than common temporal models, such as LSTM\cite{Hochreiter:1997:LSM:1246443.1246450} and GRU\cite{conf/emnlp/ChoMGBBSB14}, for the task of temporal aggregation of visual and audio features\cite{DBLP:journals/corr/MiechLS17}. One of the main drawbacks of NetVLAD is that the encoded features are in high dimension. A non-trivial classification model based on those features would need hundreds of millions of parameters. For instance, a NetVLAD network with 128 clusters will encode a feature of 2048 dimension as an vector of 262,144 dimension. A subsequent fully-connected layer with 2048-dimensional outputs will result in about 537M parameters. The parameter inefficiency would make the model harder to be optimized and easier to be overfitting.

To handle the parameter inefficiency problem, inspired by the work of ResNeXt\cite{DBLP:journals/corr/XieGDTH16}, we developed a novel neural network architecture, NeXtVLAD. Different from NetVLAD, the input features are decomposed into a group of relatively lower-dimensional vectors with attention before they are encoded and aggregated over time. The underlying assumption is that one video frame may contain multiple objects and decomposing the frame-level features before encoding would be beneficial for models to produce a more concise video representation. Experimental results on Youtube-8M dataset have demonstrated that our proposed model is more effective and efficient on parameters than the original NetVLAD model. Moreover, the NeXtVLAD model can converge faster and more resistant to overfitting.

\section{Related Works}
In this section, we provide a brief review of most relevant researches on feature aggregation and video classification.

\subsection{Feature Aggregation for Compact Video Representation}
Before the era of deep neural networks, researchers have proposed many encoding methods, including BoW (Bag of visual Words)\cite{Sivic03}, FV (Fisher Vector)\cite{conf/cvpr/PerronninD07} and VLAD (Vector of Locally Aggregated Descriptors)\cite{conf/cvpr/JegouDSP10} etc.,  to aggregate local image descriptors into a global compact vector, aiming to achieve more compact image representation and improve the performance of large-scale visual recognition. Such aggregation methods are also applied to the researches of large-scale video classification in some early works\cite{Laptev08learningrealistic}\cite{Schuldt:2004:RHA:1018429.1020906}. Recently, \cite{Arandjelovic16} proposed a differentiable module, NetVLAD, to integrate VLAD into current neural networks and achieved significant improvement for the task of place recognition. The architecture was then proved to very effective in aggregating spatial and temporal information for compact video representation\cite{DBLP:journals/corr/MiechLS17}\cite{Girdhar_17a_ActionVLAD}.

\subsection{Deep Neural Networks for Large-Scale Video Classification}
Recently, with the availability of large-scale video datasets\cite{KarpathyCVPR14}\cite{Heilbron_2015_CVPR}\cite{45619} and mass computation power of GPUs, deep neural networks have achieved remarkable advances in the field of large-scale video classification\cite{Baccouche:2011:SDL:2177908.2177914}\cite{Simonyan:2014:TCN:2968826.2968890}\cite{Ji:2013:CNN:2412386.2412939}\cite{Tran:2015:LSF:2919332.2919929}. These approaches can be roughly assigned into four categories:
(a) {\bf Spatiotemporal Convolutional Networks}\cite{KarpathyCVPR14}\cite{Ji:2013:CNN:2412386.2412939}\cite{Tran:2015:LSF:2919332.2919929}, which mainly rely on convolution and pooling to aggregate temporal information along with spatial information.
(b) {\bf Two Stream Networks}\cite{Simonyan:2014:TCN:2968826.2968890}\cite{DBLP:journals/corr/FeichtenhoferPZ16}\cite{DBLP:journals/corr/WuJWYXW15}\cite{43793}, which utilize stacked optical flow to recognize human motions in addition to the context frame images.
(c) {\bf Recurrent Spatial Networks}\cite{Baccouche:2011:SDL:2177908.2177914}\cite{DBLP:journals/corr/BallasYPC15}, which applies Recurrent Neural Networks, including LSTM or GRU to model temporal information in videos.
(d) {\bf Other approaches}\cite{Fernando_2015_CVPR}\cite{Wang_Transformation}\cite{DBLP:journals/corr/BilenFGV16}\cite{46699}, which use other solutions to generate compact features for video representation and classification.

\section{Network Architecture for NeXtVLAD}
We will first review the NetVLAD aggregation model before we dive into the details of our proposed NeXtVLAD model for feature aggregation and video classification.

\subsection{NetVLAD Aggregation Network for Video Classification}
\begin{figure}
\centering
\includegraphics[width=0.8\linewidth]{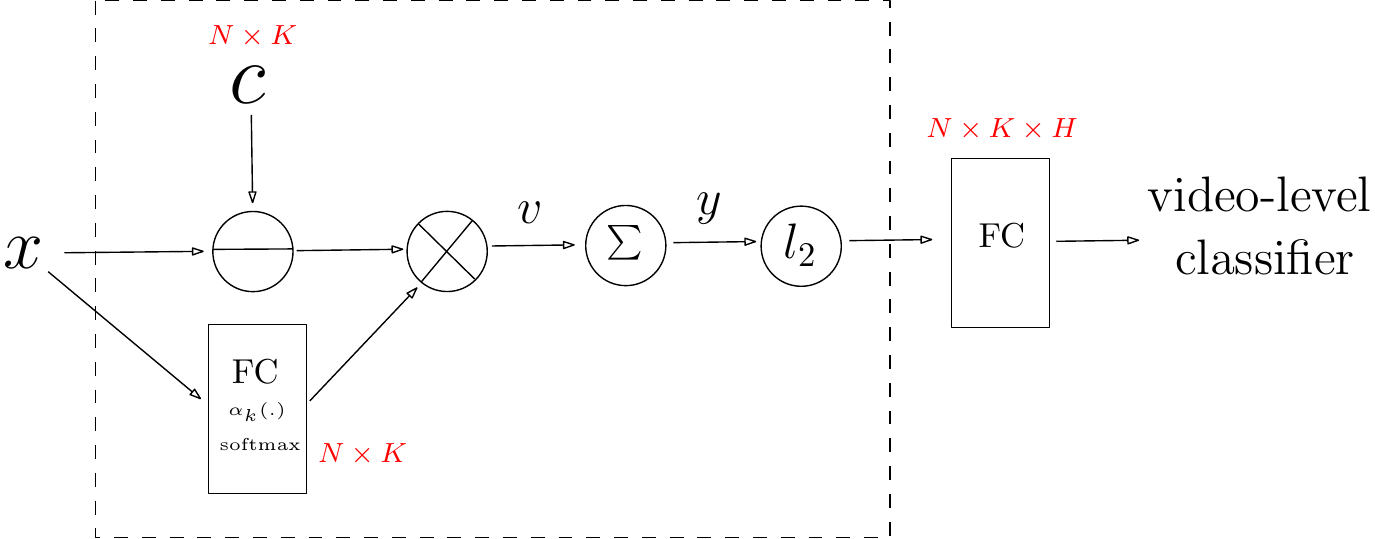}
\caption{Schema of NetVLAD model for video classification. Formulas in red denote the number of parameters (ignoring biases or batch normalization). FC means fully-connected layer.}
\label{fig:netvlad}
\end{figure}
Considering a video with $M$ frames, $N$-dimensional frame-level descriptors $x$ are extracted by a pre-trained CNN recursively. In NetVLAD aggregation of $K$ clusters, each frame-level descriptor is firstly encoded to be a feature vector of $N \times K$ dimension using the following equation:
\begin{equation}
\begin{gathered}
v_{ijk} = \alpha_k(x_i)(x_{ij} - c_{kj}) \\
i \in \{1, ..., M\}, j \in \{1, ..., N\}, k \in \{1, ..., K\}
\end{gathered}
\end{equation}
where $c_{k}$ is the $N$-dimensional anchor point of cluster $k$ and $\alpha_k(x_i)$ is a soft assignment function of $x_i$ to cluster $k$, which measures the proximity of $x_i$ and cluster $k$. The proximity function is modeled using a single fully-connected layer with softmax activation,
\begin{equation}
\alpha_k(x_i) = \frac{e^{w_k ^T x_i + b_k}}{\sum_{s=1}^K e^{w_s^T x_i + b_s}}.
\end{equation}
Secondly, a video-level descriptor $y$ can be obtained by aggregating all the frame-level features,
\begin{equation}
y_{jk} = \sum_i^M v_{ijk}
\end{equation}
and intra-normalization is applied to suppress bursts\cite{Arandjelovic:2013:VLA:2514950.2516209}.
Finally, the constructed video-level descriptor $y$ is reduced to an $H$-dimensional hidden vector via a fully-connected layer before being fed into the final video-level classifier.

As shown in Figure \ref{fig:netvlad}, the parameter number of NetVLAD model before video-level classification is about
\begin{equation}
N\times K\times (H+2),
\end{equation}
where the dimension reduction layer (second fully-connected layer) accounts for the majority of total parameters. For instance, a NetVLAD model with $N=1024$, $K=128$ and $H=2048$ contains more than $268M$ parameters.

\subsection{NeXtVLAD Aggregation Network}
\begin{figure}
\centering
\includegraphics[width=0.9\linewidth]{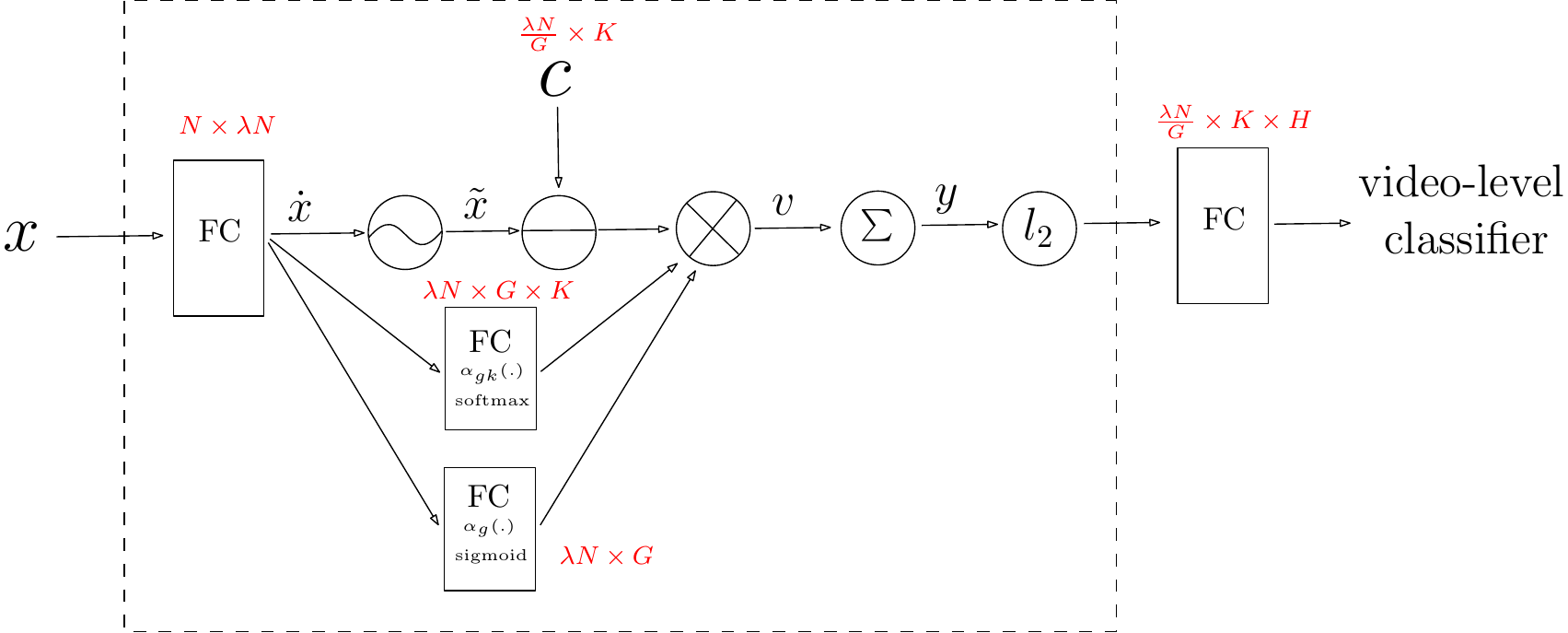}
\caption{Schema of our NeXtVLAD network for video classification. Formulas in red denote the number of parameters (ignoring biases or batch normalization). FC represents a fully-connected layer. The wave operation means a reshape transformation.}
\label{fig:nextvlad}
\end{figure}
In our NeXtVLAD aggregation network, the input vector $x_i$ is first expanded as $\dot{x}_i$ with a dimension of $\lambda N$ via a linear fully-connected layer, where $\lambda$ is a width multiplier and it is set to be 2 in all of our experiments. Then a reshape operation is applied to transform $\dot{x}$ with a shape of $(M, \lambda N)$ to $\tilde{x}$ with a shape of $(M, G, \lambda N / G)$, in which $G$ is the size of groups. The process is equivalent to splitting $\dot{x}_i$ into $G$ lower-dimensional feature vectors $\Big\{\tilde{x}^g_i \Big| g \in \{1,...,G\}\Big\}$, each of which is subsequently represented as a mixture of residuals from cluster anchor points $c_k$ in the same lower-dimensional space:
\begin{equation}
\begin{gathered}
v_{ijk}^g = \alpha_g(\dot{x}_i)\alpha_{gk}(\dot{x}_i)(\tilde{x}_{ij}^g - c_{kj})\\
g \in\{1,...,G\}, i \in \{1,...,M\}, j \in \{1,...,\frac{\lambda N}{G}\}, k \in \{1, ..., K\},
\end{gathered}
\end{equation}
where the proximity measurement of the decomposed vector $\tilde{x}_i^g$ consists of two parts for the cluster $k$:
\begin{equation}
\alpha_{gk}(\dot{x}_i) = \frac{e^{w_{gk}^T\dot{x}_i + b_{gk}}}{\sum_{s=1}^K e^{w_{gs}^T\dot{x}_i + b_{gs}}},
\end{equation}
\begin{equation}
\alpha_{g}(\dot{x}_i) = \sigma(w_g^T\dot{x}_i + b_g),
\end{equation}
in which $\sigma(.)$ is a sigmoid function with output scale from 0 to 1.
The first part $\alpha_{gk}(\dot{x}_i)$ measures the soft assignment of $\tilde{x}^g_i$ to the cluster k, while the second part $\alpha_{g}(\dot{x}_i)$ can be regarded as an attention function over groups.

Then, a video-level descriptor is achieved via aggregating the encoded vectors over time and groups:
\begin{equation}
y_{jk} = \sum_{i, g} v^g_{ijk},
\end{equation}
after which we apply an intra-normalization operation, a dimension reduction fully-connected layer and a video-level classifier as same as those of the NetVLAD aggregation network.

As noted in Figure \ref{fig:nextvlad}, because the dimension of video-level descriptors $y_{jk}$ is reduced by $G$ times compared to NetVLAD, the number of parameters shrinks. Specifically, the total number of parameters is:
\begin{equation}
\lambda N ( N + G + K(G + \frac{H + 1}{G})).
\end{equation}
Since $G$ is much smaller than $H$ and $N$, roughly speaking, the number of parameters of NeXtVLAD is about $\frac{G}{\lambda}$ times smaller than that of NetVLAD. For instance, a NeXtVLAD network with $\lambda=2$, $G=8$, $N=1024$, $K=128$ and $H=2048$ only contains $71M+$ parameters, which is about 4 times smaller than that of NetVLAD, $268M+$.

\subsection{NeXtVLAD Model and SE Context Gating}
\begin{figure}
\centering
\includegraphics[width=1.0\linewidth]{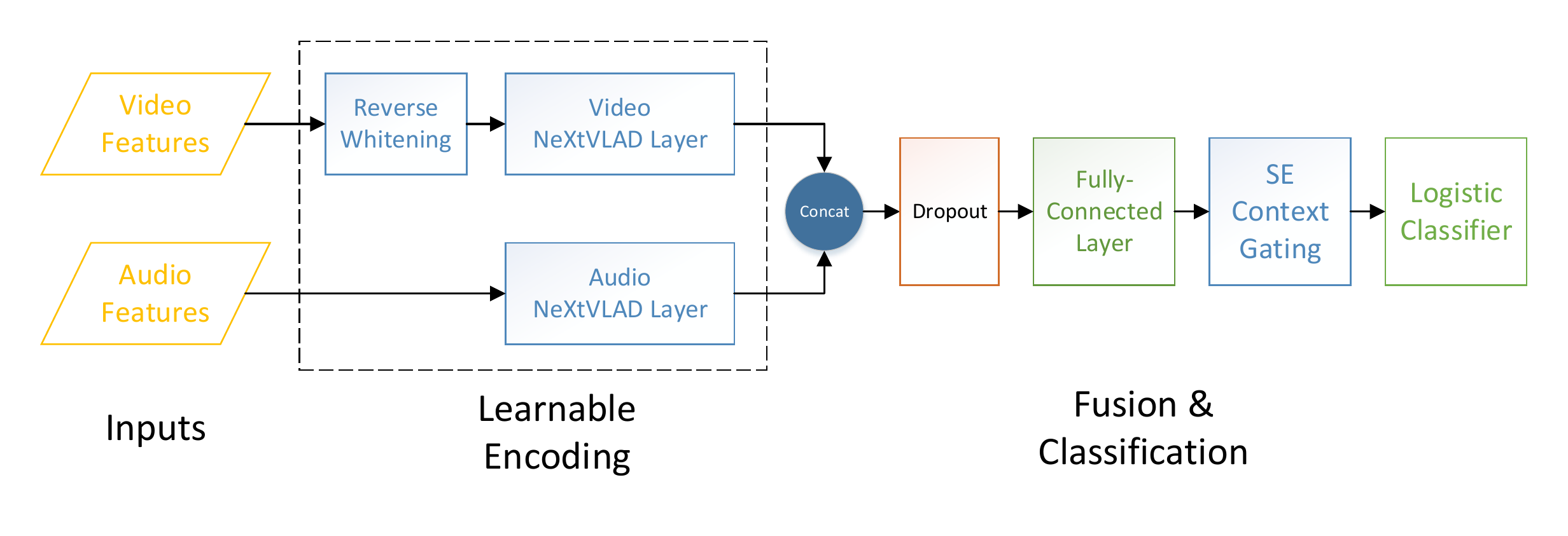}
\caption{Overview of our NeXtVLAD model designed for Youtube-8M video classification.}
\label{fig:network}
\end{figure}
The basic model we used for 2nd Youtube-8M challenge has the similar architecture with the winner solution\cite{DBLP:journals/corr/MiechLS17} for the first Youtube-8M challenge. Video and audio features are encoded and aggregated separately with a two-stream architecture. The aggregated representation is enhanced by a SE Context Gating module, aiming to modeling the dependency among labels. At last, a logistic classifier with sigmoid activation is adopted for video-level multi-label classification.

\begin{figure}
\centering
\includegraphics[width=0.8\linewidth]{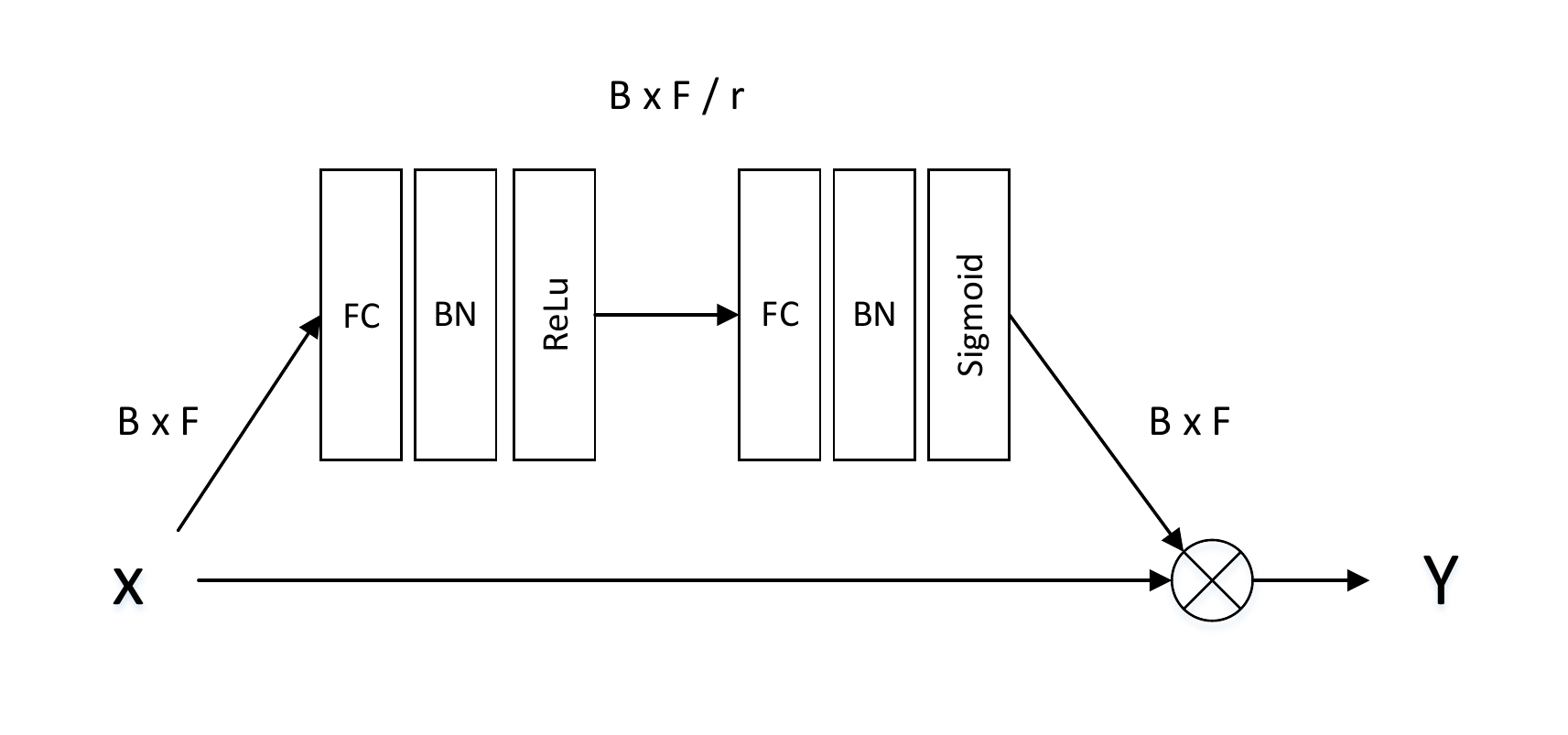}
\caption{The schema of the SE Context Gating. FC denotes fully-connected and BN denotes batch normalization. $B$ represents the batch size and $F$ means the feature size of $x$.}
\label{fig:secg}
\end{figure}

Inspired by the work of Squeeze-and-Excitation networks\cite{hu2018senet}, as shown in Figure \ref{fig:secg}, the SE Context Gating consists of 2 fully-connected layers with less parameters than the original Context Gating introduced in \cite{DBLP:journals/corr/MiechLS17}. The total number of parameters is:
\begin{equation}
\frac{2F^2}{r}
\end{equation}
where $r$ denotes the reduction ratio that is set to be 8 or 16 in our experiments.
During the competition, we find that reversing the whitening process, which is applied after performing PCA dimensionality reduction of frame-level features, is beneficial for the generalization performance of NeXtVLAD model. The possible reason is that whitening after PCA will distort the feature space by eliminating different contributions between feature dimensions with regard to distance measurements, which could be critical for the encoder to find better anchor points and soft assignments for each input feature. Since the Eigen values $\big\{e_j\big | j \in \{1,...,N\}\big\}$ of PCA transformation is released by the Google team, we are able to reverse the whitening process by:
\begin{equation}
\hat{x}_j = x_j * \sqrt{e_j}
\end{equation}
where $x$ and $\hat{x}$ are the input and reversed vector respectively.

\subsection{Knowledge Distillation with On-the-fly  Naive Ensemble}
Knowledge distillation\cite{44873}\cite{DBLP:journals/corr/ZhangXHL17}\cite{DBLP:journals/corr/LiH16e} was designed to transfer the generalization ability of the cumbersome teacher model to a relatively simpler student network by using prediction from teacher model as an additional ``soft target'' during training. During the competition, we tried the network architecture introduced in \cite{on_the_fly} to distill knowledge from a on-the-fly mixture prediction to each sub-model.
\begin{figure}
\centering
\includegraphics[width=0.9\linewidth]{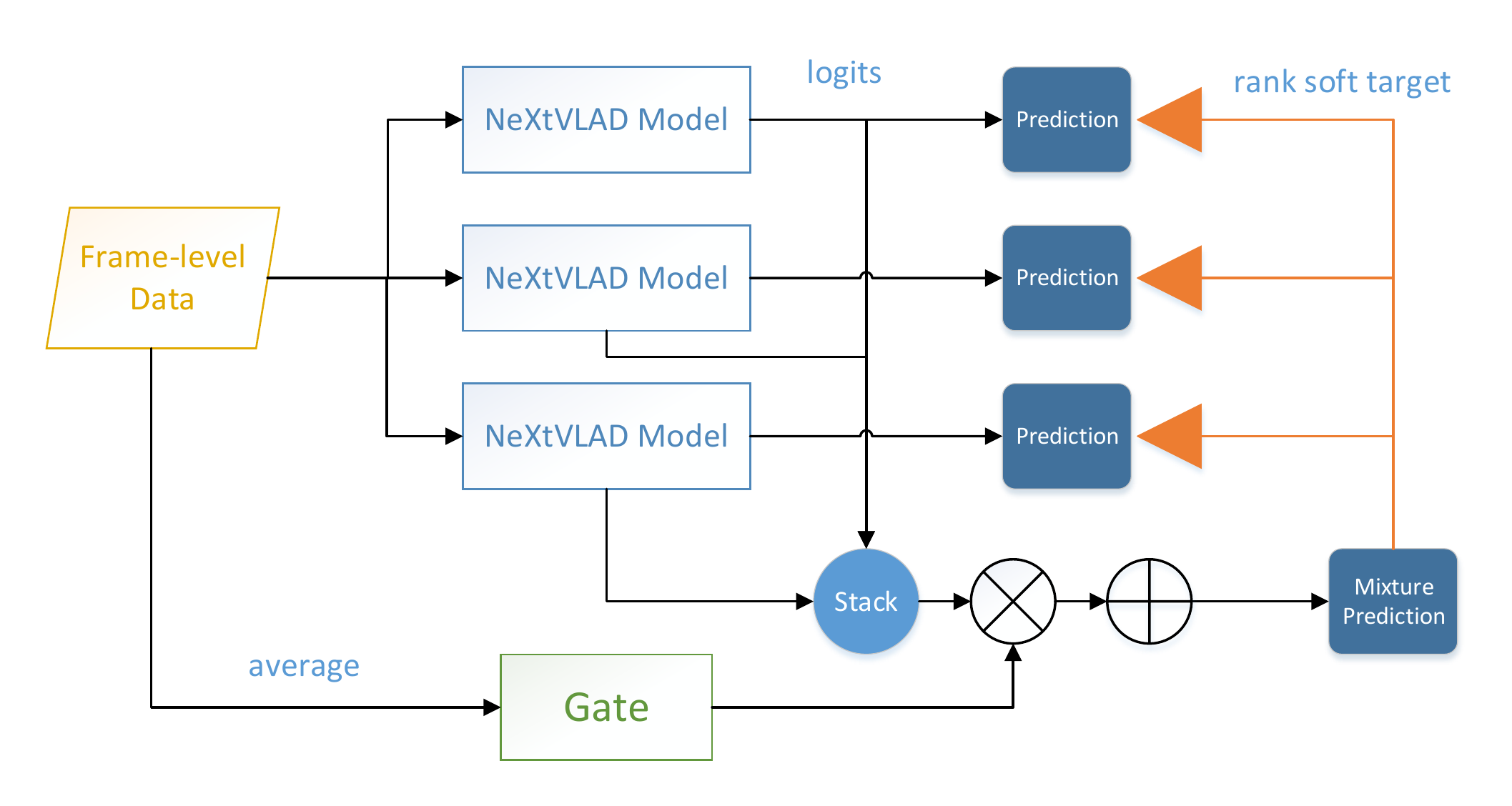}
\caption{Overview of a mixture of 3 NeXtVLAD models with on-the-fly knowledge distillation. The orange arrows indicate the distillation of knowledge from mixture predictions to the sub-models.}
\label{fig:network}
\end{figure}

As shown in Figure \ref{fig:network}, the logits of the mixture predictions $z^e$ is a weighted sum of logits $\big\{ z^m \big | m \in \{1,2,3\} \big\}$ from the 3 corresponding sub-models:
\begin{equation}
z^e = \sum_{m=1}^3 a_m(\bar{x}) * z^m
\end{equation}
where $a_m(.)$ represents the gating network,
\begin{equation}
a_m(\bar{x}) = \frac{e^{w_m^T\bar{x} + b_m}}{\sum_s^3 e^{w_s^T \bar{x} + b_s}}
\end{equation}
and $\bar{x}$ represents the frame mean of input features $x$. The knowledge of the mixture prediction is distilled to each sub-model through minimizing the KL divergence written as:
\begin{equation}
\mathcal{L}_{kl}^{m, e} = \sum_{c=1}^C p^e(c) \log\frac{p^e(c)}{p^m(c)},
\end{equation}
where C is the total number of class labels and $p(.)$ represents the rank soft prediction:
\begin{equation}
p^m(c) = \frac{exp(z^m_c / T)}{\sum_{s=1}^C exp(z^m_s / T)}, \ p^e(c) = \frac{exp(z^e_c / T)}{\sum_{s=1}^C exp(z^e_s / T)}.
\end{equation}
where $T$ is a temperature which can adjust the relative importance of logits.
As suggested in \cite{44873}, larger $T$ will increase the importance of logits with smaller values and encourage models to share more knowledge about the learned similarity measurements of the task space.
The final loss of the model is:
\begin{equation}
\mathcal{L} = \sum_{m=1}^3 \mathcal{L}_{bce}^{m} + \mathcal{L}_{bce}^e + T^2 * \sum_{m=1}^3\mathcal{L}_{kl}^{m, e}
\end{equation}
where $\mathcal{L}_{bce}^m$ ($\mathcal{L}_{bce}^e$) means the binary cross entropy between the ground truth labels and prediction from model $m$ (mixture prediction).

\section{Experimental Results}
This section provides the implementation details and presents our experimental results on the Youtube-8M dataset\cite{45619}.

\subsection{Youtube-8M Dataset}
Youtube-8M dataset (2018) consists of about 6.1M videos from Youtube.com, each of which has at least 1000 views with video time ranging from 120 to 300 seconds and is labeled with one or multiple tags (labels) from a vocabulary of 3862 visual entities. These videos further split into 3 partitions: train(70\%), validate(20\%) and test(10\%). Along with the video ids and labels, visual and audio features are provided for every second of the videos, which are referred as frame-level features. The visual features consists of hidden representations immediately prior to the classification layer in Inception\cite{Ioffe:2015:BNA:3045118.3045167}, which is pre-trained on Imagenet\cite{Deng09imagenet:a}. The audio features are extracted from a audio classification CNN\cite{DBLP:journals/corr/HersheyCEGJMPPS16}. PCA and whitening are then applied to reduce the dimension of visual and audio feature to 1024 and 128 respectively.

In the 2nd Youtube-8M video understanding challenge, submissions are evaluated using Global Average Precision(GAP) at 20. For each video, the predictions are sorted by confidence and the GAP score is calculated as:
\begin{equation}
GAP = \sum_{i=1}^{20} p(i)*r(i)
\end{equation}
in which $p(i)$ is the precision and $r(i)$ is the recall given the top $i$ predictions.

\subsection{Implementation Details}
Our implementation is based on the TensorFlow\cite{Abadi:2016:TSL:3026877.3026899} starter code\footnote{https://github.com/google/youtube-8m}. All of the models are trained using the Adam optimizer\cite{DBLP:journals/corr/KingmaB14} with an initial learning rate of 0.0002 on two Nvidia 1080 TI GPUs. The batch size is set to be 160 (80 on each GPU). We apply a $l_2$(1e-5) regularizer to the parameters of the video-level classifier and use a dropout ratio of 0.5 aiming to avoid overfitting. No data augmentation is used in training NeXtVLAD models and the padding frames are masked out during the aggregation process via:
\begin{equation}
v_{ijk}^g = mask(i)\alpha_g(\dot{x}_i)\alpha_{gk}(\dot{x}_i)(\tilde{x}_{ij}^g - c_{kj})
\end{equation}
where
\begin{equation}
mask(i) =
  \begin{cases}
    1       & \quad \text{if } \ \ i <= M\\
    0  & \quad \text{else}
  \end{cases}
\end{equation}

In all the local experiments, models are trained for 5 epochs (about 120k steps) using only the training partition and the learning rate is exponentially decreased by a factor of 0.8 every 2M samples. Then the model is evaluated using only about $\frac{1}{10}$ of the evaluation partition, which is consistently about 0.002 smaller than the score at public leaderboard\footnote{https://www.kaggle.com/c/youtube8m-2018/leaderboard} for the same models. As for the final submission model, it is trained for 15 epochs (about 460k steps) using both training and validation partitions and the learning rate is exponentially decreased by a factor of 0.9 every 2.5M samples. More details can be found at \url{https://github.com/linrongc/youtube-8m}.

\subsection{Model Evaluation}
\begin{table}
\caption{Performance (on local validation partition) comparison for single aggregation models. The parameters inside parenthesis represents (group number $G$, dropout ratio, cluster number $K$, hidden size $H$)}\label{single_compare}
  \centering
  \begin{tabular}{lcc}
    \hline
    Model & Parameter & GAP \\
    \hline
    NetVLAD (-, 0.5drop, 128K, 2048H) & 297M & 0.8474\\
    NetVLAD\_random (-, 0.5drop, 256K, 1024H) & 274M & 0.8507\\
    NetVLAD\_small (-, 0.5drop, 128K, 2048H) & 88M & 0.8582 \\
    \hline
    NeXtVLAD (32G, 0.2drop, 128K, 2048H) & 55M & 0.8681\\
    NeXtVLAD (16G, 0.2drop, 128K, 2048H) & 58M & 0.8685\\
    NeXtVLAD (16G, 0.5drop, 128K, 2048H) & 58M & 0.8697\\
    NeXtVLAD (8G, 0.5drop, 128K, 2048H) & 89M & 0.8723\\
    \hline
  \end{tabular}
\end{table}
We evaluate the performance and parameter efficiency of individual aggregation models in Table \ref{single_compare}. For fair comparison, we apply a reverse whitening layer for video features, a dropout layer after concatenation of video and audio features and a logistic model as the video-level classifier in all the presented models. Except for NetVLAD\_random which sampled 300 random frames for each video, all the other models didn't use any data augmentation techniques. NetVLAD\_small use a linear fully-connected layer to reduce the dimension of inputs to $\frac{1}{4}$ of the original size for visual and audio features, so that the number of parameters are much comparable to other NeXtVLAD models.
\begin{figure}
\centering
\includegraphics[width=0.9\linewidth]{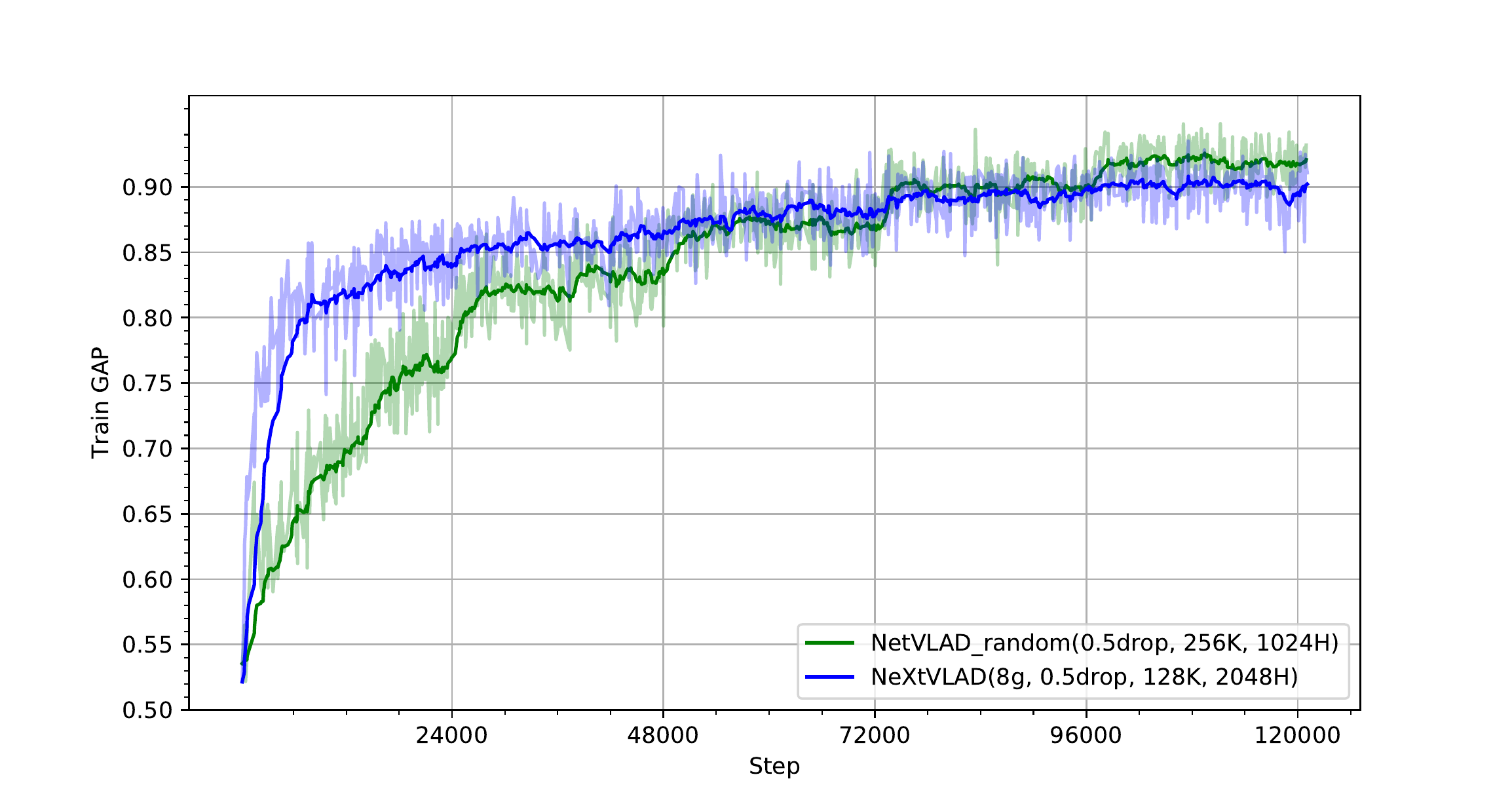}
\caption{Training GAP on Youtube-8M dataset. The ticks of x axis are near the end of each epoch.}
\label{fig:train}
\end{figure}

From Table \ref{single_compare}, one can observe that our proposed NeXtVLAD neural networks are more effective and efficient on parameters than the original NetVLAD model by a significantly large margin. With only about $30\%$ of the size of NetVLAD\_random model\cite{DBLP:journals/corr/MiechLS17}, NeXtVLAD increase the GAP score by about 0.02, which is a significant improvement considering the large size of Youtub-8M dataset. Furthermore, as shown in Figure \ref{fig:train}, the NeXtVLAD model is converging faster, which reaches a training GAP score of about 0.85 in just 1 epoch.

Surprisingly, the NetVLAD model performs even worse than the NetVLAD\_small model, which indicates NetVLAD models tend to overfit the training dataset. Another interesting observation in Figure \ref{fig:train} is that the most of GAP score gains happens around the beginning of a new epoch for NetVLAD model. The observation implies that the NetVLAD model are more prone to remember the data instead of find useful feature patterns for generalization.
\begin{table}
\caption{The GAP scores of submissions during the competition. All the other parameters used are (0.5drop, 112K, 2048H). The final submissions are tagged with * }\label{final_submission}
  \centering
  \begin{tabular}{lccc}
    \hline
    Model & Parameter & Private GAP & Public GAP \\
    \hline
    single NeXtVLAD(460k steps) & 79M & 0.87846 & 0.87910\\
    3 NeXtVLAD (3T, 250k steps) & 237M & 0.88583 & 0.88657\\
    3 NeXtVLAD (3T, 346k steps) & 237M & 0.88681 & 0.88749\\
    3 NeXtVLAD* (3T, 460k steps) & 237M & 0.88722 & 0.88794\\
    3 NeXtVLAD* (3T, 647k steps) & 237M & 0.88721 & 0.88792\\
    \hline
  \end{tabular}\\
\end{table}

To meet the competition requirements, we use an ensemble of 3 NeXtVLAD models with parameters (0.5drop, 112K, 2048H), whose size is about 944M bytes. As shown in Table \ref{final_submission}, training longer can always lead to better performance of NeXtVLAD models. Our best submission is trained about 15 epochs, which takes about 3 days on two 1080 TI GPUs. If we only retain one branch from the mixture model, a single NeXtVLAD model with only 79M parameters will achieve a GAP score of 0.87846, which could be ranked 15/394 in the final leaderboard.

Due to time and resource limit, we set the parameters $T=3$, which is the temperature in on-the-fly knowledge distillation, as suggested in \cite{on_the_fly}. We ran an AB test experiments after the competition, as shown in \ref{abtest}, somehow indicates $T=3$ is not optimal. Further tuning of the parameter could result in a better GAP score.
\begin{table}
\caption{The results (on local validation set) of an AB test experiment for $T$ tuning.}\label{abtest}
  \centering
  \begin{tabular}{lccc}
    \hline
    Model & Parameter & local GAP \\
    \hline
    3 NeXtVLAD (0T, 120k steps) & 237M & 0.8798\\
    3 NeXtVLAD (3T, 120k steps) & 237M & 0.8788\\
    \hline
  \end{tabular}\\
\end{table}

\section{Conclusion}
In this paper, a novel NeXtVLAD model is developed to support large-scale video classification under budget constraints. Our NeXtVLAD model has provided a fast and efficient network architecture to aggregate frame-level features into a compact feature vector for video classification. The experimental results on Youtube-8M dataset have demonstrated that our proposed NeXtVLAD model is more effective and efficient on parameters than the previous NetVLAD model, which is the winner of the first Youtube-8M video understanding challenge.

\section{Acknowledgement}
The authors would like to thank Kaggle and the Google team for hosting the Youtube-8M video understanding challenge and providing the Youtube-8M Tensorflow Starter Code.
\bibliographystyle{splncs}
\bibliography{egbib}
\end{document}